\documentclass[10pt,twocolumn,letterpaper]{article}
\usepackage{subcaption}
\usepackage{graphicx,booktabs,makecell,array}
\usepackage{array}  
\usepackage{tabularx} 
\newcolumntype{Y}{>{\centering\arraybackslash}X}
\usepackage[pagenumbers]{wacv} 
\usepackage{multirow}

\usepackage{microtype}

\definecolor{wacvblue}{rgb}{0.21,0.49,0.74}
\usepackage[pagebackref,breaklinks,colorlinks,allcolors=wacvblue]{hyperref}
\def\wacvPaperID{2237} 
\def\confName{WACV}
\def\confYear{2026}

\title{CSGaussian: Progressive Rate-Distortion Compression and Segmentation for\\ 3D Gaussian Splatting} 
\author{
Yu-Jen Tseng$^{1}$,
Chia-Hao Kao$^{2}$,
Jing-Zhong Chen$^{1}$,
Alessandro Gnutti$^{2}$,\\[4pt]
Shao-Yuan Lo$^{3}$,
Yen-Yu Lin$^{1}$,
Wen-Hsiao Peng$^{1}$\\[4pt]
$^1$National Yang Ming Chiao Tung University, Taiwan\\
$^2$University of Brescia, Italy \quad 
$^3$National Taiwan University, Taiwan\\[4pt]
\tt\small alan.cs12@nycu.edu.tw, 
wpeng@cs.nycu.edu.tw\\
}

\begin{document}

\twocolumn[{%
\renewcommand\twocolumn[1][]{#1}%
\maketitle

\includegraphics[width=1\linewidth]{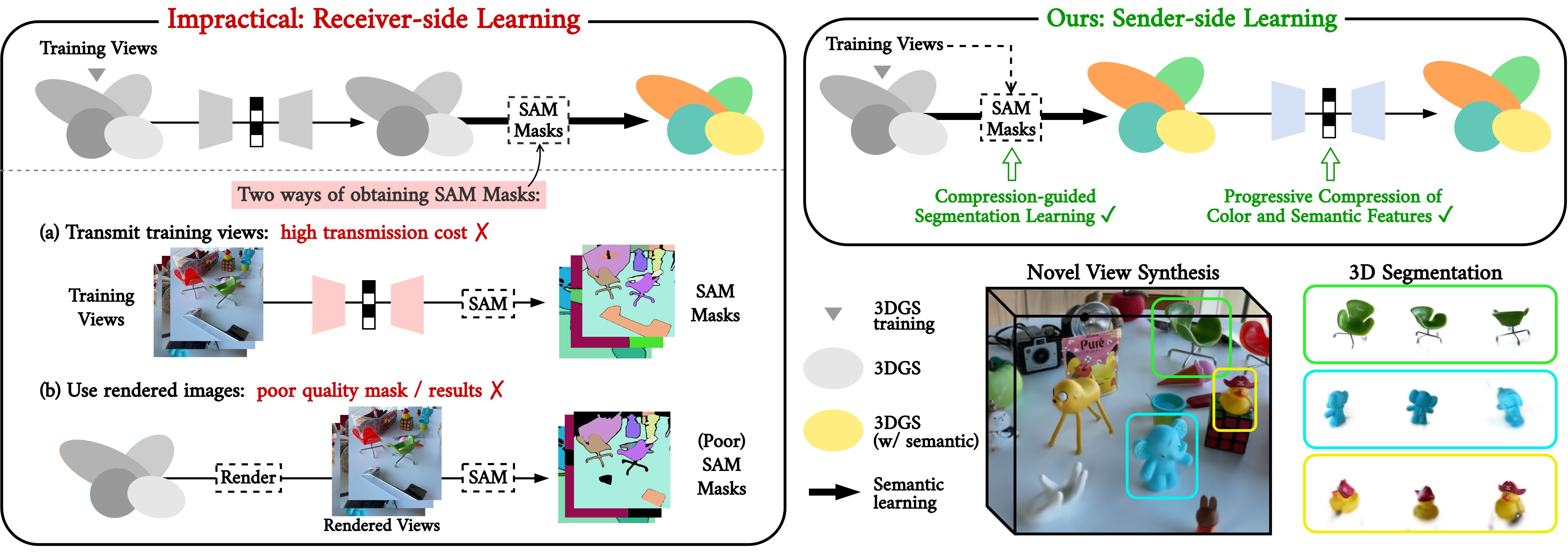}

\captionof{figure}{
Comparison of approaches for efficient 3DGS segmentation at the receiver. Left: The naive solution of learning semantics at the receiver is impractical, as it either (a) requires transmitting all training-view images, or (b) relies on rendered views that yield poor-quality SAM masks and suboptimal segmentation. Right: Our proposed method introduces sender-side compression-guided segmentation learning and transmits the semantically enriched 3DGS with RD-optimized compression, enabling efficient 3D segmentation at the receiver.
}
\vspace{2em}
  \label{fig:teaser}
}]

\begin{abstract}
We present the first unified framework for rate-distortion-optimized compression and segmentation of 3D Gaussian Splatting (3DGS). While 3DGS has proven effective for both real-time rendering and semantic scene understanding, prior works have largely treated these tasks independently, leaving their joint consideration unexplored. Inspired by recent advances in rate-distortion-optimized 3DGS compression, this work integrates semantic learning into the compression pipeline to support decoder-side applications--such as scene editing and manipulation--that extend beyond traditional scene reconstruction and view synthesis. Our scheme features a lightweight implicit neural representation-based hyperprior, enabling efficient entropy coding of both color and semantic attributes while avoiding costly grid-based hyperprior as seen in many prior works. To facilitate compression and segmentation, we further develop compression-guided segmentation learning, consisting of quantization-aware training to enhance feature separability and a quality-aware weighting mechanism to suppress unreliable Gaussian primitives. Extensive experiments on the LERF and 3D-OVS datasets demonstrate that our approach significantly reduces transmission cost while preserving high rendering quality and strong segmentation performance.
\end{abstract}
    
\vspace{-1em}
\section{Introduction}
\label{sec:intro}
\newcommand{\commentout}[1]{} 

Recent advances in 3D vision have made 3D Gaussian Splatting (3DGS)~\cite{kerbl20233dgaussiansplattingrealtime} a promising representation for 3D scenes due to its excellent reconstruction quality and real-time rendering capabilities. However, 3DGS typically involves a large number of Gaussian primitives, which leads to high storage cost and memory footprint. This limitation has spurred research into efficient 3DGS compression. 

Aimed at minimizing parameter count, memory footprint, and file size, early research focuses primarily on developing more compact 3DGS representations~\cite{lu2023scaffoldgsstructured3dgaussians, lee2024compact3dgaussianrepresentation, fan2024lightgaussianunbounded3dgaussian}.
%
More recently, rate-distortion (RD)-optimized compression of 3DGS~\cite{chen2024hachashgridassistedcontext, zhan2025cat3dgscontextadaptivetriplaneapproach, wang2024contextgscompact3dgaussian, wang2024end} emerged as a new school of thought, targeting transmission and storage efficiency. Departing from earlier approaches that merely reduce parameter count without entropy coding, these methods perform quantization and entropy encoding of Gaussian attributes to achieve a balanced trade-off between compressed file size (\textit{rate}) and rendering quality (\textit{distortion}) via end-to-end, per-scene optimization.

Meanwhile, the rising demand for 3D scene understanding in applications such as robotics and autonomous driving has catalyzed rapid advancements in 3DGS segmentation~\cite{wu2024opengaussian, li2025instancegaussian, lyu2024gaga, cen2025segment, qin2024langsplat, shi2024language, ye2024gaussian}. 
Recent methods learn 3D semantic features by leveraging multi-view images and their associated camera poses. A common approach to learning semantic features involves self-supervised contrastive learning guided by foundation models, such as Segment Anything Model (SAM)~\cite{kirillov2023segment} or CLIP~\cite{radford2021learning}. Conceptually, the semantic features of individual Gaussian primitives are acquired by first rendering them onto 2D training views and then aligning them with 2D segmentation masks produced by SAM or other foundation models. These semantic features empower the resulting 3DGS representation to support downstream tasks, including open-vocabulary segmentation~\cite{liu2023weakly, qin2024langsplat, kerr2023lerf} and 3D scene manipulation~\cite{chen2024gaussianeditor, huang20253d, wu2025aurafusion360, xiao2025localized}.

This work explores a novel application scenario in which 3DGS optimization is performed on a server, while rendering, scene editing, manipulation, and understanding are carried out on a remote end device—such as an Augmented Reality (AR) headset. A naive approach would transmit color-only Gaussian primitives from the server (sender) and attempt to learn semantic features locally on the headset (receiver). However, this solution is infeasible as it either requires additionally sending all training-view images, which is prohibitively expensive (Figure~\ref{fig:teaser}a), or learning semantics from rendered views of compressed 3DGS, which leads to severe performance degradation (Figure~\ref{fig:teaser}b). In the latter case, the compromised visual quality also undermines the effectiveness of pre-trained vision models, such as SAM, leading to poor mask generation and suboptimal segmentation outcomes. A more practical solution is to learn semantic features on the server and transmit them alongside other color-related Gaussian attributes, as illustrated on the right of Figure~\ref{fig:teaser}. This approach calls for an efficient compression system that accounts for both color and semantic features.

So far, only few studies~\cite{li2025instancegaussian, dai2025efficient} have explored 3DGS compression and segmentation simultaneously. InstanceGS~\cite{li2025instancegaussian} utilizes a hierarchical structure~\cite{lu2023scaffoldgsstructured3dgaussians} to efficiently represent both color and semantic features, while DF-3DGS~\cite{dai2025efficient} decouples the semantic field from color information. Both approaches work under the assumption that semantic features require less granularity than color and are amenable to compression. Nevertheless, they do not consider rate-distortion (RD)-optimized compression of color and segmentation features. 
%


\begin{figure}[t]
    \centering
    \resizebox{\linewidth}{!}{
    \includegraphics[width=0.45\textwidth]{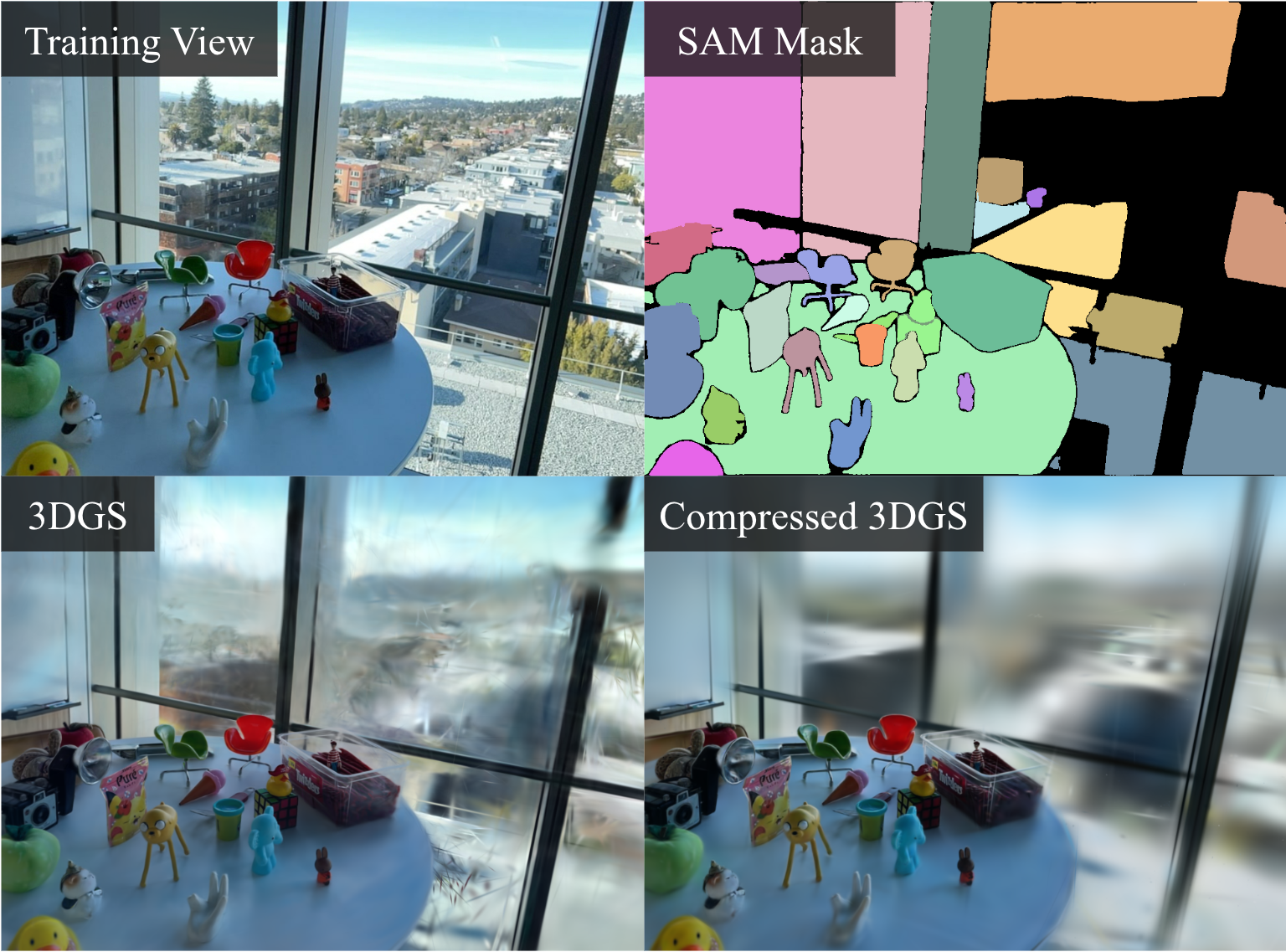}  
    }
    \caption{
    Visualization of training view, SAM mask, and 3DGS renderings with and without compression. 
    Background regions of 3DGS, especially when compressed, contain low-quality primitives that fail to represent objects meaningfully. This discrepancy with SAM masks hinders segmentation learning when all primitives are treated equally.
    }
    \label{fig:observation}
    \vspace{-3mm}
\end{figure}

In this paper, we present the first 3DGS framework to perform progressive RD-optimized compression and segmentation, enabling efficient yet accurate representation and understanding of 3D scenes. Our proposed scheme, CSGaussian, employs a hierarchical anchor-based representation~\cite{lu2023scaffoldgsstructured3dgaussians} to encode both color and semantic information. To facilitate their entropy coding, we incorporate an Implicit Neural Representation (INR)-based hyperprior to model their distributions. To support compression and segmentation, we propose a compression-guided segmentation learning strategy that tightly couples the two tasks via  (1) quantization-aware training and (2) quality-aware weighting. The former is applied to semantic features; it improves both compression and segmentation performance. The latter stems from the observation that Gaussian primitives vary in quality, as shown in Figure~\ref{fig:observation}.
Low-quality and background primitives often misalign with semantic masks, thereby hindering contrastive learning of semantic features when they are treated equally.
%
Compared to RD-optimized 3DGS frameworks that compress only color~\cite{chen2024hachashgridassistedcontext, zhan2025cat3dgscontextadaptivetriplaneapproach, wang2024contextgscompact3dgaussian}, our method progressively encodes color and semantic information into a single bitstream, supporting both high-fidelity reconstruction and accurate segmentation.
   
In summary, this work makes three primary contributions:
\begin{itemize}
    \item We pioneer a novel RD-optimized framework for simultaneous compression and segmentation on 3DGS.    
    \item Our framework introduces a simple yet effective INR-based hyperprior for Gaussian attributes along with semantic information, leading to a significant bitrate reduction.
    \item We develop a quantization-aware training strategy and a quality-aware weighting mechanism to notably improve 3D scene understanding.
\end{itemize}
Extensive experiments demonstrate the effectiveness of our method, which substantially reduces transmission cost while performing favorably against prior methods.

\section{Related Work}
\label{sec:related_work}

\subsection{3DGS Compression}

3DGS~\cite{kerbl20233dgaussiansplattingrealtime} represents a 3D scene using numerous learnable Gaussian primitives. High-quality, real-time rendering requires substantial storage and bandwidth, highlighting the need for efficient compression.
%

\vspace{-0.15in}
\paragraph{Compact 3DGS Representation.}

Early efforts on 3DGS compression primarily focus on making the representation more compact in terms of storage size and memory footprint. One representative paradigm is to directly reduce the number of Gaussian primitives by pruning those that contribute minimally to rendered image quality~\cite{lee2024compact3dgaussianrepresentation, fan2024lightgaussianunbounded3dgaussian, girish2024eaglesefficientaccelerated3d, ali2024trimmingfatefficientcompression}. 
Another strategy employs vector quantization to quantize Gaussian attributes into fixed-length codewords in a learned codebook~\cite{fan2024lightgaussianunbounded3dgaussian, lee2024compact3dgaussianrepresentation, navaneet2024compgssmallerfastergaussian, niedermayr2023compressed}.
Several existing methods explore hierarchical representations of Gaussian primitives~\cite{lu2023scaffoldgsstructured3dgaussians, ren2024octreegsconsistentrealtimerendering, sun2024f3dgsfactorizedcoordinatesrepresentations}. 
For example, Scaffold-GS~\cite{lu2023scaffoldgsstructured3dgaussians} employs an anchor-based design, where a group of Gaussian primitives is represented by one anchor, significantly reducing the parameter count.


\vspace{-0.15in}
\paragraph{Rate-distortion-optimized 3DGS Representation.}

Building on recent advances in learned image and video compression~\cite{ballé2018variationalimagecompressionscale, he2022elicefficientlearnedimage, balle2017end}, several recent works have started integrating entropy coding into 3DGS frameworks to enable end-to-end RD-optimization~\cite{chen2025hac++, chen2024hachashgridassistedcontext, zhan2025cat3dgscontextadaptivetriplaneapproach, zhan2025cat3dgspronewbenchmark, wang2024contextgscompact3dgaussian, liu2025hemgshybridentropymodel, liu2024compgsefficient3dscene}. These methods typically apply scalar quantization to Gaussian attributes and model their distributions with a hyperprior, enabling efficient arithmetic coding. For instance, HAC~\cite{chen2024hachashgridassistedcontext}, built on Scaffold-GS, presents one of the earliest RD-optimized 3DGS compression systems by leveraging a hash grid to capture spatial relationships. CAT-3DGS~\cite{zhan2025cat3dgscontextadaptivetriplaneapproach} further improves coding efficiency through a triplane-based hyperprior and autoregressive modeling in both spatial and channel dimensions. Similarly, ContextGS~\cite{wang2024contextgscompact3dgaussian} exploits local correlations between Gaussian primitives to construct an autoregressive model for entropy coding their attributes. Although these methods achieve a good trade-off between transmission bitrate and rendering quality, they do not incorporate semantic information or support 3D scene understanding.



\subsection{3DGS Segmentation}

Integrating 3DGS with vision foundation models~\cite{kirillov2023segment, radford2021learning, caron2021emerging, lilanguage}, has led to significant progress in 3DGS segmentation~\cite{ wu2024opengaussian, peng2025d, qin2024langsplat, cen2025segment, choi2024click, li2025instancegaussian, zhangeconsg}. 
%
The core principle involves attaching semantic features to Gaussian primitives, which allows 3D scenes to be represented by both color and semantic information.
Based on the way of feature acquisition, existing methods fall into two categories: feature distillation and mask lifting. 
%


\vspace{-0.15in}
\paragraph{Feature Distillation.}

These approaches distill features from foundation models and embed them into 3DGS. 
For instance, LangSplat~\cite{qin2024langsplat} employs an autoencoder to project CLIP features into a 3D latent space, while LEGaussians~\cite{shi2024language} applies codebook quantization to distill CLIP and DINO~\cite {caron2021emerging} features. 
On the other hand, Feature3DGS~\cite{zhou2024feature} extracts features from the SAM encoder and employs the SAM decoder for segmentation. 
However, this early line of work has a key drawback: it requires a dedicated decoder to recover features during inference, leading to high computational overhead and inefficient rendering.


\vspace{-0.15in}
\paragraph{Mask Lifting.}

Another line of research explored learning 3D semantic features by leveraging segmentation masks produced by vision foundation models through self-supervised learning objectives. 
%
%
For example, SAGA~\cite{cen2025segment} adopts a classic contrastive loss with SAM masks to guide the learning process.
ClickGaussian~\cite{choi2024click} introduces global feature-guided learning to address cross-view inconsistency.
More recently, OpenGaussian~\cite{wu2024opengaussian} combines SAM masks and CLIP embeddings and introduces a 3D-to-2D feature association for open-vocabulary segmentation. 
Building on this, InstanceGS~\cite{li2025instancegaussian} reformulates the representation within the Scaffold-GS framework, enabling Gaussian primitives associated with the same anchor to share semantic features.
%
Compared to feature distillation, mask lifting offers a more scalable and computationally efficient solution by circumventing the need to decode dense feature embeddings.

However, most existing methods for 3DGS segmentation overlook the storage and transmission costs associated with 3DGS representations and their learned semantic features, which poses a challenge in practical applications.

\section{Preliminary}

\subsection{Scaffold-GS: Anchor-based Representation}

Scaffold-GS ~\cite{lu2023scaffoldgsstructured3dgaussians} introduces a storage-efficient, anchor-based representation for organizing Gaussian primitives. Specifically, anchor points are initialized on a predefined voxel grid, and each anchor carries information about a fixed number of $K$ (e.g. 10) Gaussian primitives. For example, the positions of these Gaussian primitives, denoted as $\{\boldsymbol{\mu}_i \in \mathbb{R}^{3} \}_{i=1}^{K}$, are computed from the anchor position $\boldsymbol{x}$, learnable offsets $\{\boldsymbol{O}_i \in \mathbb{R}^{3}\}_{i=1}^{K}$ and a scaling factor $\boldsymbol{l} \in \mathbb{R}^{6}$, following the formulation: \( \{\boldsymbol{\mu}_i\}_{i=1}^K = \boldsymbol{x} + \{\boldsymbol{O}_i\}_{i=1}^K \cdot \boldsymbol{l} \). Additional structural and color attributes of the Gaussian primitives, including the color $\{\boldsymbol{c}_i \in \mathbb{R}^{3} \}_{i=1}^{K}$, opacity $\{\boldsymbol{\alpha}_i \in \mathbb{R}^{1}\}_{i=1}^{K}$, scale $\{\boldsymbol{S}_i \in \mathbb{R}^{3}\}_{i=1}^{K}$, and rotation $\{\boldsymbol{R}_i \in \mathbb{R}^{4}\}_{i=1}^{K}$, are decoded from the anchor feature $\boldsymbol{f} \in \mathbb{R}^{50}$ via an MLP decoder.





\subsection{Contrastive Semantic Feature Learning}
\label{subsec:prel_feattures_learning}

In mask lifting-based 3DGS segmentation frameworks--such as OpenGaussian~\cite{wu2024opengaussian} and InstanceGS~\cite{li2025instancegaussian}--a semantic feature \( \boldsymbol{s} \in \mathbb{R}^{6} \) is learned for each Gaussian primitive to encode its semantic meaning. 
Similar to color attributes, these features can be rendered onto a 2D feature map \( \boldsymbol{F} \) using $\alpha$-blending~\cite{kerbl20233dgaussiansplattingrealtime}. To guide the self-supervised learning of semantic features, 2D segmentation masks $\boldsymbol{M}$ are first generated by SAM for each training image. The semantic learning process is then supervised using two complementary losses: (1) the intra-mask smoothing loss $\mathcal{L}_s$, which encourages semantic features within the same 2D mask to converge toward their mean, and (2) the inter-mask contrastive loss $\mathcal{L}_c$, which promotes the separation between semantic features correspond to different instance masks. In symbols, we have:
\begin{equation}
\mathcal{L}_s = \sum_{i=1}^{m} \sum_{h=1}^{H} \sum_{w=1}^{W} \boldsymbol{M}_{i,h,w} \cdot \| \boldsymbol{F}_{:,h,w} - \boldsymbol{\bar{F}}_i \|^2,
\end{equation}
\begin{equation}
\mathcal{L}_c = \frac{1}{m(m-1)} \sum_{i=1}^{m} \sum_{j=1, j \neq i}^{m} \frac{1}{\| \boldsymbol{\bar{F}}_i - \boldsymbol{\bar{F}}_j \|^2},
\end{equation}
where \( H \) and \( W \) denote the height and width of the image, respectively, \( m \) is the number of SAM masks in the current view, and \(\bar{\boldsymbol{F}}_i\), \(\bar{\boldsymbol{F}}_j\) are the mean features of two distinct instance masks. 

\section{Method}

\subsection{System Overview}
\begin{figure*}
  \centering
  \begin{subfigure}{1\linewidth}
    \centering
    \includegraphics[width=\linewidth]{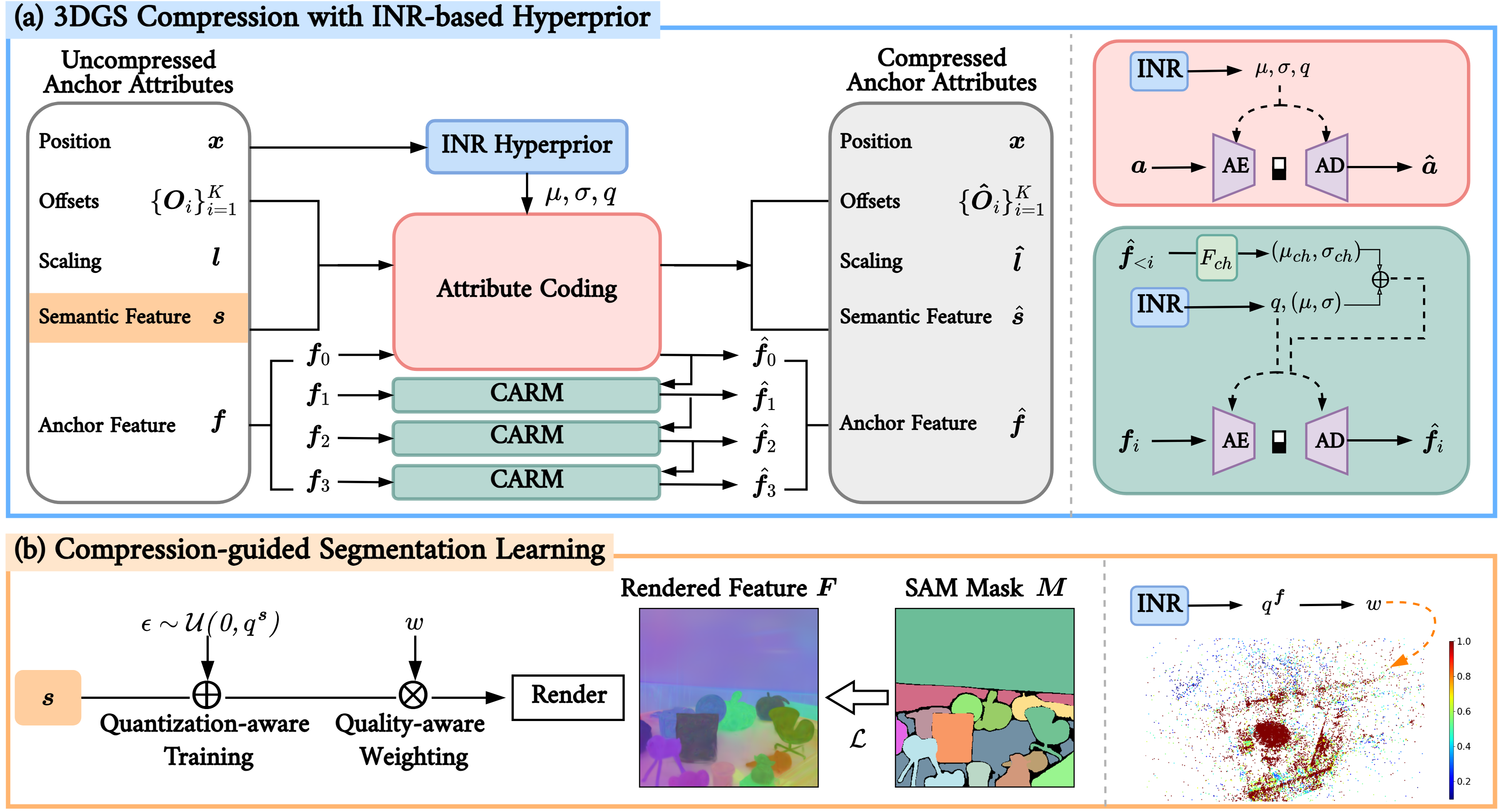}
  \end{subfigure}
  \vspace{-1.5em}
  \caption{Overview of our proposed framework. Top (a): The left illustrates the overall architecture of our compression system with an INR-based hyperprior, and the right details attribute coding for attributes \( \boldsymbol{a} \) and channel-wise autoregressive modeling (CARM). For better visualization, only one INR module is depicted, though two models are employed to compress color and semantic features. Bottom (b): Pipeline of compression-guide segmentation learning, including quantization-aware training and quality-aware weighting. The weighting visualization (bottom right) shows foreground anchors are emphasized with larger weightings while background ones are down-weighted.}
  \label{fig:overview}
\end{figure*}

In this work, we propose a novel RD-optimized 3DGS compression system that enables efficient transmission of color and semantic information of a 3D scene within a single bitstream (Figure~\ref{fig:overview}). To achieve this, we adopt Scaffold-GS~\cite{lu2023scaffoldgsstructured3dgaussians} as our base 3DGS representation and structure the encoding process into three main stages, as outlined below.
\vspace{-0.15in}
\paragraph{Stage 1: Color-only 3DGS Optimization.} The first stage focuses exclusively on learning color (appearance) attributes in an end-to-end, RD-optimized manner, targeting both rendering quality and transmission bitrate.
In a way similar to prior RD-optimized 3DGS compression methods~\cite{chen2024hachashgridassistedcontext, zhan2025cat3dgscontextadaptivetriplaneapproach, chen2025hac++}, we leverage entropy coding to efficiently encode quantized anchor attributes, including the offsets $\{{\boldsymbol{O}_i} \}^K_{i=1}$, scaling $\boldsymbol{l}$, and anchor feature $\boldsymbol{f}$. Unlike existing approaches that rely on heavy spatial priors such as hash grids or triplanes, we propose a lightweight hyperprior based on an implicit neural representation (INR). Taking the 3D position of each anchor as input, the INR hyperprior learns a function to model the probability distributions of anchor attributes scatted throughout the 3D space without requiring hash grids or triplanes. 

\vspace{-0.15in}
\paragraph{Stage 2: Compression-guided Semantic Feature Learning.} This step aims to learn semantic features $\boldsymbol{s}$ associated with every anchor, to support 3D segmentation. Since semantic information typically require less granularity than color information~\cite{dai2025efficient, li2025instancegaussian}, all Gaussian primitives associated with an anchor share the same semantic feature vector. Our feature learning builds on the self-supervised framework of~\cite{wu2024opengaussian, li2025instancegaussian}, with two key enhancements motivated by the joint demands of RD-optimized compression and segmentation: (i) quantization-aware training for semantic features, and (ii) quality-aware weighting (Sec.~\ref{par:qwm}).  

\vspace{-0.15in}
\paragraph{Stage 3: Semantic Feature Compression.} Lastly, semantic learning and semantic feature compression are jointly optimized within an end-to-end, RD-optimized framework. To entropy encode (or decode) semantic features, their probability distributions are estimated by a dedicated INR-based hyperprior model. Both the semantic features and the hyperprior are updated simultaneously to balance segmentation accuracy against compressed bitrate.

\subsection{INR-based Hyperprior}
\label{sec:method/compression}

To ensure efficient transmission of 3DGS, we propose a lightweight INR-based hyperprior. It comprises two distinct models: one for Stage~1 (color compression) and another for Stage~3 (semantic feature compression). As depicted in Figure~\ref{fig:overview}a, our INR-based hyperprior models the distributions of 3DGS anchor attributes for entropy coding. Specifically, it consists of two compact multilayer perceptrons. The first predicts the distributions of anchor features \(\boldsymbol{f}\), offsets \(\{\boldsymbol{O}_i\}_{i=1}^{K}\), and scaling factors \(\boldsymbol{l}\); the second models the distributions of semantic features \(\boldsymbol{s}\). Given the positional embedding~\cite{vaswani2017attention} of an anchor location \(\boldsymbol{x}\), the model outputs the corresponding Gaussian distribution parameters--means $\boldsymbol{\mu}$ and variances $\boldsymbol{\sigma}$--along with the quantization step size q for each attribute type. In particular, each attribute type is assigned a distinct quantization step size, which is shared across its components. The probability of a quantized attribute \(\hat{\boldsymbol{a}}=q\cdot\text{round}(\boldsymbol{a}/q)\) is then estimated as the probability mass under the learned Gaussian prior:  
\[
p(\hat{\boldsymbol{a}} \mid \boldsymbol{x}) = \int_{\hat{\boldsymbol{a}} - \frac{q}{2}}^{\hat{\boldsymbol{a}} + \frac{q}{2}} \mathcal{N}\bigl(\boldsymbol{\mu}, \boldsymbol{\sigma}\bigr)\,d\boldsymbol{a}.
\]
 Our grid-free formulation enables faster training and rendering compared to triplane-based hyperpriors, which require additional compression of grid points on each triplane typically through time-consuming autoregressive modeling. 
%

To further enhance the entropy modeling for color attributes \(\boldsymbol{f}\), we integrate a Channel-wise AutoRegressive Model (CARM) following~\cite{zhan2025cat3dgscontextadaptivetriplaneapproach, zhan2025cat3dgspronewbenchmark}. This model sequentially predicts the distribution of feature channels conditioned on previously decoded ones, thereby exploiting intra-feature dependencies. In practice, the predicted means \(\boldsymbol{\mu}_{ch}\) and variances \(\boldsymbol{\sigma}_{ch}\) are added to the initial prediction output by the proposed INR-based hyperprior. We also adopt a view-frequency-aware masking mechanism~\cite{chen2024hachashgridassistedcontext, zhan2025cat3dgscontextadaptivetriplaneapproach} to prune primitives with little contribution to rendering quality.
\subsection{Compression-guided Semantic Learning}
\label{sec:method/3d_to_4d}

We enable contrastive semantic feature learning (Sec.~\ref{subsec:prel_feattures_learning}) in Stage~2. 
The joint pursuit of compression and segmentation motivates enhancements to the feature learning process. As illustrated in Figure \ref{fig:overview}b, we introduce two key techniques: quantization-aware training and quality-aware weighting.

\vspace{-0.15in}
\paragraph{Quantization-Aware Training (QAT).}\label{par:qat}~Quantization-aware training, a widely adopted technique in learned image and video compression~\cite{balle2017end, he2022elicefficientlearnedimage, ballé2018variationalimagecompressionscale}, aims to approximate quantization effects during training and enable end-to-end optimization under a rate-distortion objective. In a similar vein, we simulate scalar quantization with step size $q^{\boldsymbol{s}}$ by injecting uniform noise $\epsilon \sim \mathcal{U}(0,q^{\boldsymbol{s}})$~\cite{balle2017end} to the learned semantic features $\boldsymbol{s}$ prior to rendering them as a 2D feature map for self-supervised learning.
%
%
\textcolor{black}{In Stage 2, $q^{\boldsymbol{s}}$ is set to a fixed value $\mathcal{Q}$ since RD-optimized compression is activated only in Stage 3. In Stage 3, $q^{\boldsymbol{s}}$ is subsequently refined by a modulation factor predicted by the INR-based hyperprior.} 

In passing, QAT serves a dual purpose in our framework. Beyond simulating quantization effects for end-to-end optimization, the additive noise has a regularization effect on semantic learning itself. Since the contrastive loss aims to enforce separation between semantic features corresponding to different SAM masks, additive noise implicitly encourages semantic features to be pushed further apart. The resulting features are more robust and remain distinguishable under perturbations or noise (See Sec.~\ref{sec:exp/ablation} and supplementary material for an in-depth analysis of this aspect). This result is consistent with the observation made in~\cite{10003114, yu2024noisynnexploringimpactinformation}.

\vspace{-0.15in}
\paragraph{Quality-aware Weighting Mechanism (QWM).}\label{par:qwm}~Existing 3DGS segmentation methods typically optimize semantic features by treating all Gaussian primitives uniformly. However, Gaussian primitives that contribute to the rendering of the background regions are often less well learned and tend to generate poor rendering quality in those regions. They can thus mislead and negatively impact the semantic learning process because SAM masks in these background regions can be inconsistent or even random. This, in turn, has a significant impact on the contrastive learning of foreground semantic features.


We address this issue through a quality-aware weighting mechanism that explicitly accounts for the reliability of each anchor during contrastive learning. Notably, the quantization step size $q^{\boldsymbol{f}}$, estimated by the INR-based hyperprior in Stage 1, serves as a strong proxy for anchor quality. Intuitively, anchors with smaller step sizes tend to be more reliable and are often situated in foreground-like regions, whereas those with larger step sizes are typically less accurate and of lower quality. This relationship establishes a natural link between compression and segmentation.


Formally, we specify a weight ${w}$ for each anchor according to $q^{\boldsymbol{f}}$ as follows:  
\begin{equation}
\label{eq:qwm}
    {w} = 
    \begin{cases}
      1, & \text{if}\ 1 - \lVert q^{\boldsymbol{f}} \rVert \geq\mathcal{T} \\
      1 - \lVert q^{\boldsymbol{f}} \rVert, & \text{otherwise,}
    \end{cases}
\end{equation}
where $\mathcal{T}$ is an empirically chosen threshold and $\lVert \cdot \rVert$ denotes a normalization operator that maps its input to an interval [0, 1]. As a result, low-quality background anchors with large step sizes are down-weighted, effectively minimizing their influence and treating them as if they carried little semantic content (see the bottom-right of Figure~\ref{fig:overview}b). With both QAT and QWM, the modulated semantic features rendered onto a 2D image plane for contrastive learning is \(\tilde{\boldsymbol{s}} = (\boldsymbol{s} + \epsilon) \cdot w\).

\begin{figure*}



     \centering
      \begin{subfigure}[t]{\columnwidth}
          \centering
          \includegraphics[width=\linewidth]{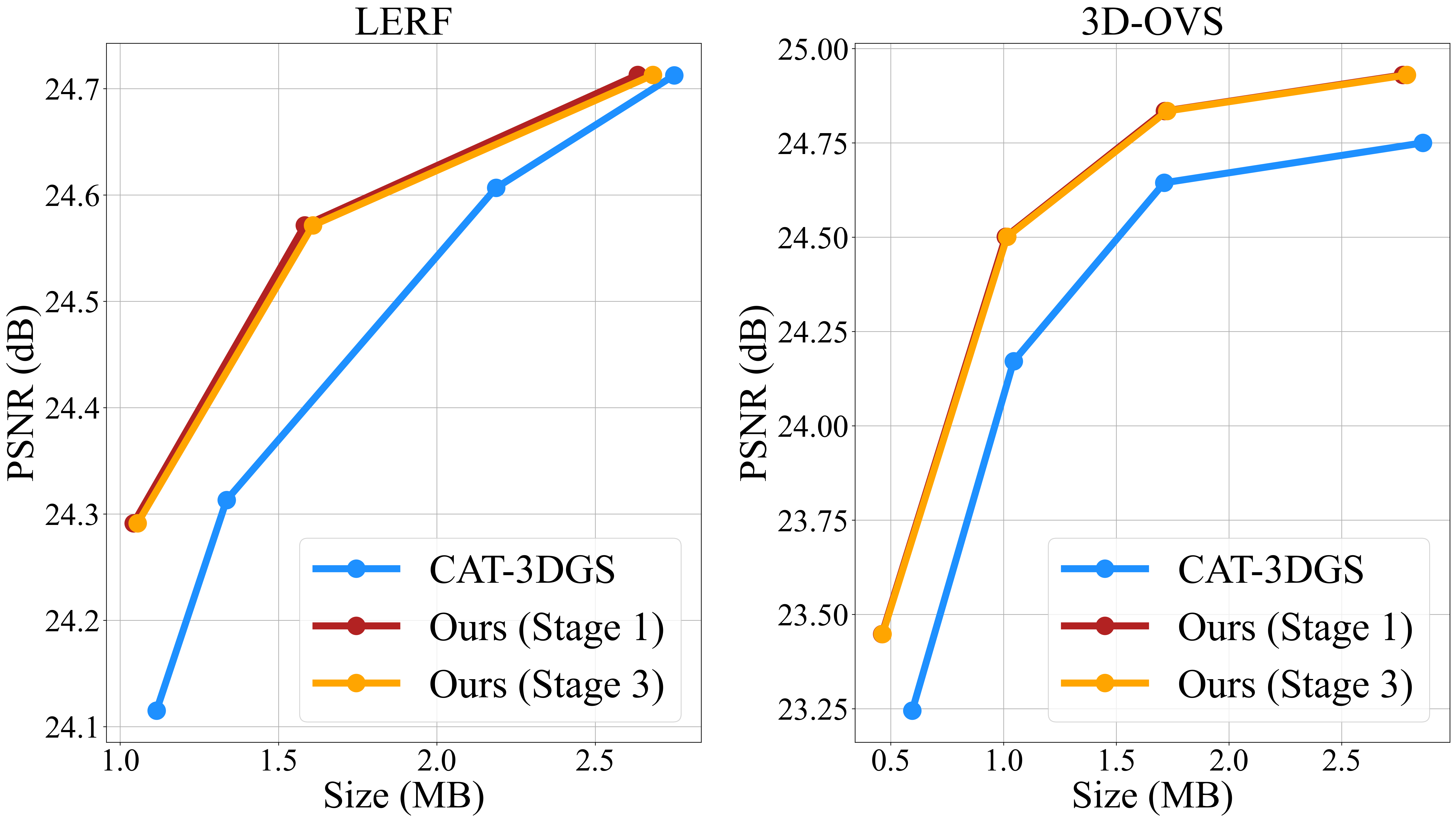}
          \caption{Reconstruction Quality}  
          \label{fig:rd_psnr}
      \end{subfigure}
      \begin{subfigure}[t]{\columnwidth}
          \centering
          \includegraphics[width=\linewidth]{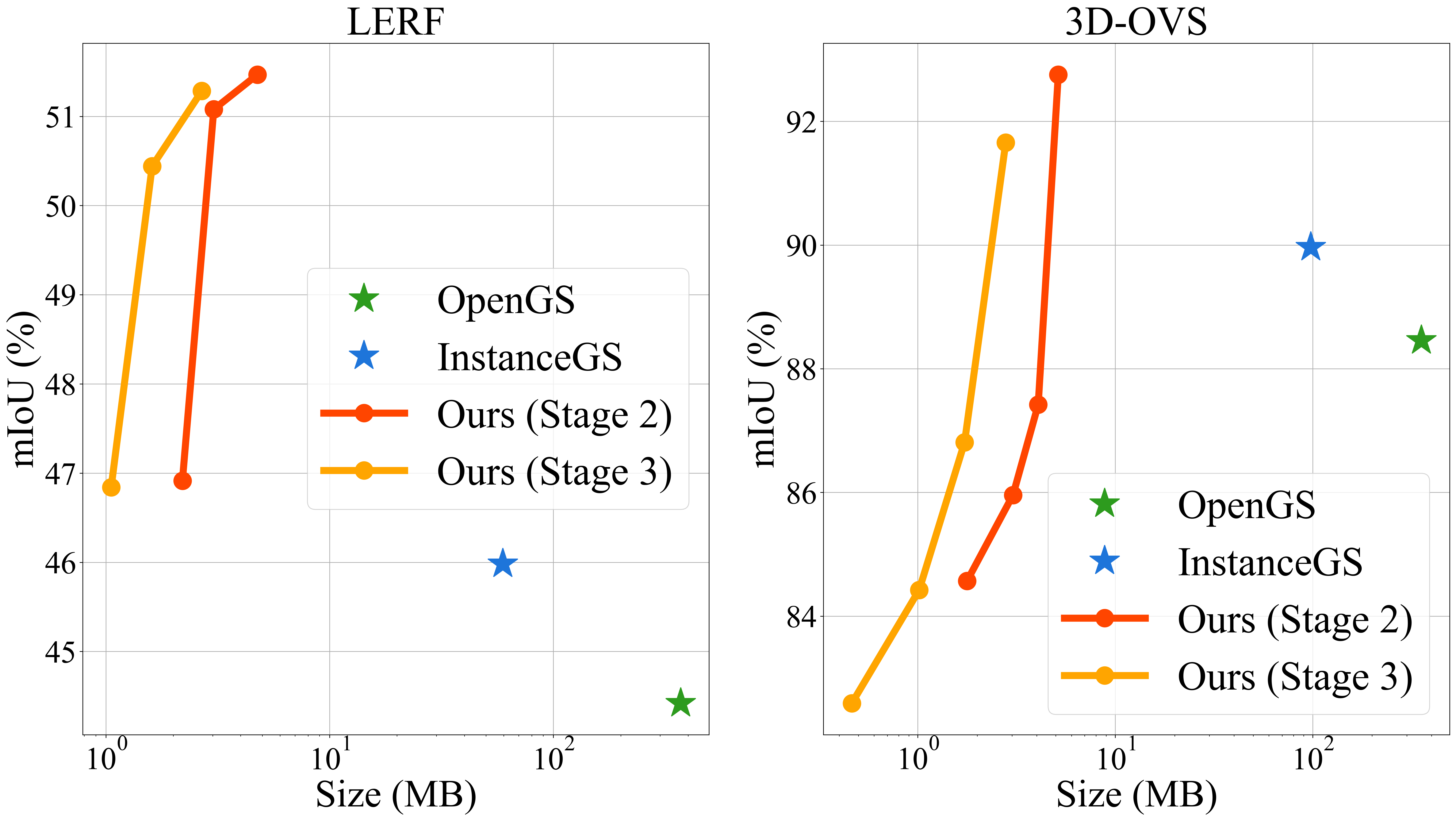}
          \caption{Segmentation Performance} 
          \label{fig:rd_miou}
      \end{subfigure}
      \vspace{-0.5em}
      \caption{Performance comparison, where the x axis represents bitrate, and y axis represents rendering quality or segmentation accuracy.}
      \label{fig:main}
    
\end{figure*}

\subsection{Training Objectives}
\label{sec:training_objectives}

Our framework addresses color/appearance learning, semantic learning, and semantic feature compression in three progressive stages. Each stage has a specific training objective. 

Stage 1 learns a 3DGS representation to capture color information of the 3D scene in an RD-optimized manner by a typical rate-distortion training objective~\cite{chen2024hachashgridassistedcontext, zhan2025cat3dgscontextadaptivetriplaneapproach}:
%
\[
\mathcal{L}_{\text{Stage 1}} = \mathcal{L}_{\text{distortion}} + \lambda_{\text{rate}} \left( \mathcal{L}_{\text{rate}} + \lambda_{\text{offset mask}}\, \mathcal{L}_{\text{offset mask}} \right),
\]
where \( \mathcal{L}_{\text{distortion}} \) measures rendering quality using a combination of L1 loss and SSIM, and \( \mathcal{L}_{\text{rate}} \) denotes the average number of bits required to entropy encode an anchor. The term \( \mathcal{L}_{\text{offset mask}} \) promotes spatial sparsity by penalizing redundant offsets. The trade-offs among these factors are governed by the weighting factors \( \lambda_{\text{rate}} \) and \( \lambda_{\text{offset mask}} \). Notably, \( \mathcal{L}_{\text{offset mask}} \) is multiplied further by \( \lambda_{\text{rate}} \) to enable more aggressive pruning of anchors at lower bitrates.  

With the frozen anchor features, position, scaling, and offsets, Stage 2 introduces semantic features to each anchor and optimizes them for 3D segmentation. We adopt the self-supervised learning scheme from~\cite{wu2024opengaussian, li2025instancegaussian}, guided by SAM-generated masks. The training objective $\mathcal{L}_{\text{Stage 2}}$ includes both the inter-mask contrastive loss $\mathcal{L}_c$ and intra-mask smoothing loss $\mathcal{L}_s$ in Sec.~\ref{subsec:prel_feattures_learning}. In this stage, no entropy coding is applied. 

In the final stage, semantic features are entropy coded using a separate INR hyperprior model. Here, the semantic features and the hyperprior are jointly optimized with another rate-distortion objective that incorporates the estimated bitrate $\mathcal{L}_{\text{feature rate}}$ of semantic features in addition to the feature loss $\mathcal{L}_{\text{Stage 2}}$:  
\[
\mathcal{L}_{\text{Stage 3}} = \mathcal{L}_{\text{Stage 2}} + \lambda_{\text{feature rate}} \ \mathcal{L}_{\text{feature rate}},
\]
where \( \lambda_{\text{feature rate}} \) controls the trade-off between the two terms. Our progressive training optimizes semantic features to minimize transmission overhead while maximizing segmentation quality, without compromising rendering fidelity.

\section{Experiments}
\label{sec:exp/experiments}
\subsection{Experimental Setup}

\paragraph*{\textbf{Datasets and Metrics.}}
We evaluate our proposed 3DGS compression and segmentation framework on two representative datasets: LERF~\cite{kerr2023lerf} and 3D-OVS~\cite{liu2023weakly}. 
The LERF dataset, designed for 3D object localization, contains complex real-world indoor scenes captured with an iPhone and is annotated by~\cite{shi2024language}. 
In contrast, the 3D-OVS dataset focuses on forward-facing scenes targeting open-vocabulary 3D segmentation. 
Following the evaluation protocols adopted in~\cite{qin2024langsplat, wu2024opengaussian, li2025instancegaussian}, we evaluate open-vocabulary 3D segmentation with mean Intersection-over-Union (mIoU) and Peak Signal-to-Noise Ratio (PSNR) as metrics for segmentation accuracy and reconstruction quality, respectively. 
To assess coding efficiency, we also report the compressed file size (bitrate) in megabytes (MB).


\vspace{-0.15in}
\paragraph*{\textbf{Implementation Details.}}

We implement our method based on \cite{lu2023scaffoldgsstructured3dgaussians, zhan2025cat3dgscontextadaptivetriplaneapproach} using the PyTorch framework and train it on a single NVIDIA RTX 4090 GPU. 
During quantization-aware training in Stage 2, we set the quantization step size \( \mathcal{Q} \) to 1 and the threshold \( \mathcal{T} \) of quality-aware weighting to 0.65. 
To evaluate performance across different bitrates, we use a range of rate-control factors \(\lambda_{\text{rate}}\). 
Specifically, we choose \(\lambda_{\text{rate}} = \{0.004, 0.01, 0.02\}\) for the LERF dataset and \(\lambda_{\text{rate}} = \{0.004, 0.01, 0.02, 0.05\}\) for the 3D-OVS dataset.


\vspace{-0.15in}
\paragraph*{\textbf{Baselines.}}  
Our comparison encompasses two key aspects: reconstruction quality and segmentation performance, both of which also consider transmission costs. 
For reconstruction, we assess the efficiency of our system with INR-based hyperprior against CAT-3DGS~\cite{zhan2025cat3dgscontextadaptivetriplaneapproach} with a triplane hyperprior. 
We directly quantify the overhead of enabling segmentation at the decoder side by evaluating 3DGS after both Stage 1 (color-only compression) and Stage 3 (with semantic features) of our framework. 
For segmentation, we compare our method against two recent state-of-the-art 3DGS segmentation methods, OpenGaussian~\cite{wu2024opengaussian} and InstanceGS~\cite{li2025instancegaussian}, which do not incorporate RD-optimized compression. 
To further examine the impact of semantic feature compression, we include Stage 2 results, where semantic features are optimized but left uncompressed, alongside the final Stage 3 results of our method.

\subsection{Performance Comparison}
\label{sec:exp/main_rd}

\paragraph*{\textbf{Reconstruction Quality.}}
Figure~\ref{fig:rd_psnr} shows that our proposed framework achieves 3D segmentation with marginal bitrate overhead (2-4\%) relative to the color-only 3DGS representation (Ours, Stage 1), without compromising reconstruction quality. 
Compared to CAT-3DGS~\cite{zhan2025cat3dgscontextadaptivetriplaneapproach}, which relies on a parameter-heavy triplane, our INR-based hyperprior achieves superior RD performance without the need to learn and signal triplanes.
As shown in Table~\ref{tab:time}, our method further reduces average training and decoding times of CAT-3DGS by 30\% and 50\%, respectively, while maintaining high rendering speed on the LERF dataset. This confirms the superiority of our proposed method in terms of both coding efficiency and computational cost.

\vspace{-0.15in}
\paragraph*{\textbf{Segmentation Performance.}}

As shown in Figure~\ref{fig:rd_miou}, our framework delivers substantial performance gains over state-of-the-art 3D Gaussian segmentation methods, which do not incorporate RD-optimized compression. Compared to OpenGaussian~\cite{wu2024opengaussian} and InstanceGS~\cite{li2025instancegaussian}, our method achieves bitrate reductions of over 140× and 23×, respectively, through efficient entropy coding. Simultaneously, it improves mIoU by 2–5\% via compression-guided segmentation learning.


Focusing on semantic feature compression (Stage 2 to Stage 3), our INR-based hyperprior consistently demonstrates high efficiency in compressing additional semantic features—significantly reducing transmission cost while preserving segmentation performance. The key distinction between Stage 2 and Stage 3 lies in their treatment of semantic features: Stage 2 learns uncompressed features, whereas Stage 3 introduces an INR-based hyperprior to encode them in a RD-optimized manner.


\commentout{our INR-based hyperprior consistently outperforms both the factorized prior and the uncompressed baseline}

\subsection{Ablation Studies}
\label{sec:exp/ablation}

We conduct several ablation studies to validate the effectiveness of our proposed compression-guided learning and to justify key hyperparameter choices. All the experiments are conducted on the LERF~\cite{kerr2023lerf} dataset. 
\textcolor{black}{Except for the first one, all experiments adopt the same compressed color-only 3DGS (Stage 1) and keep the additional semantic features uncompressed to ensure fair comparisons.}

\begin{figure*}[t]
    \centering
    
    \begin{subfigure}[t]{0.24\textwidth}
        \centering
        \includegraphics[width=\linewidth]{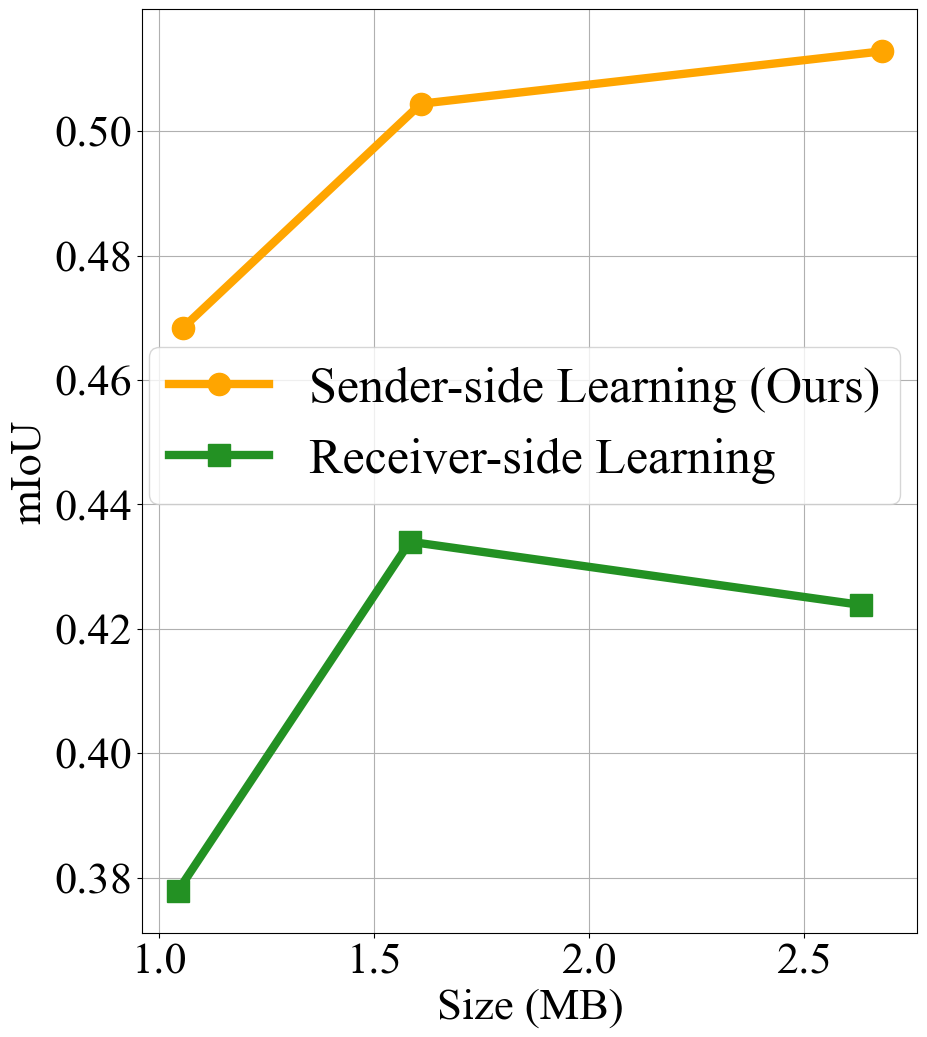}
        \caption{Receiver-side Learning}
        \label{fig:rendered}
    \end{subfigure}
    \hfill
    \begin{subfigure}[t]{0.24\textwidth}
        \centering
        \includegraphics[width=\linewidth]{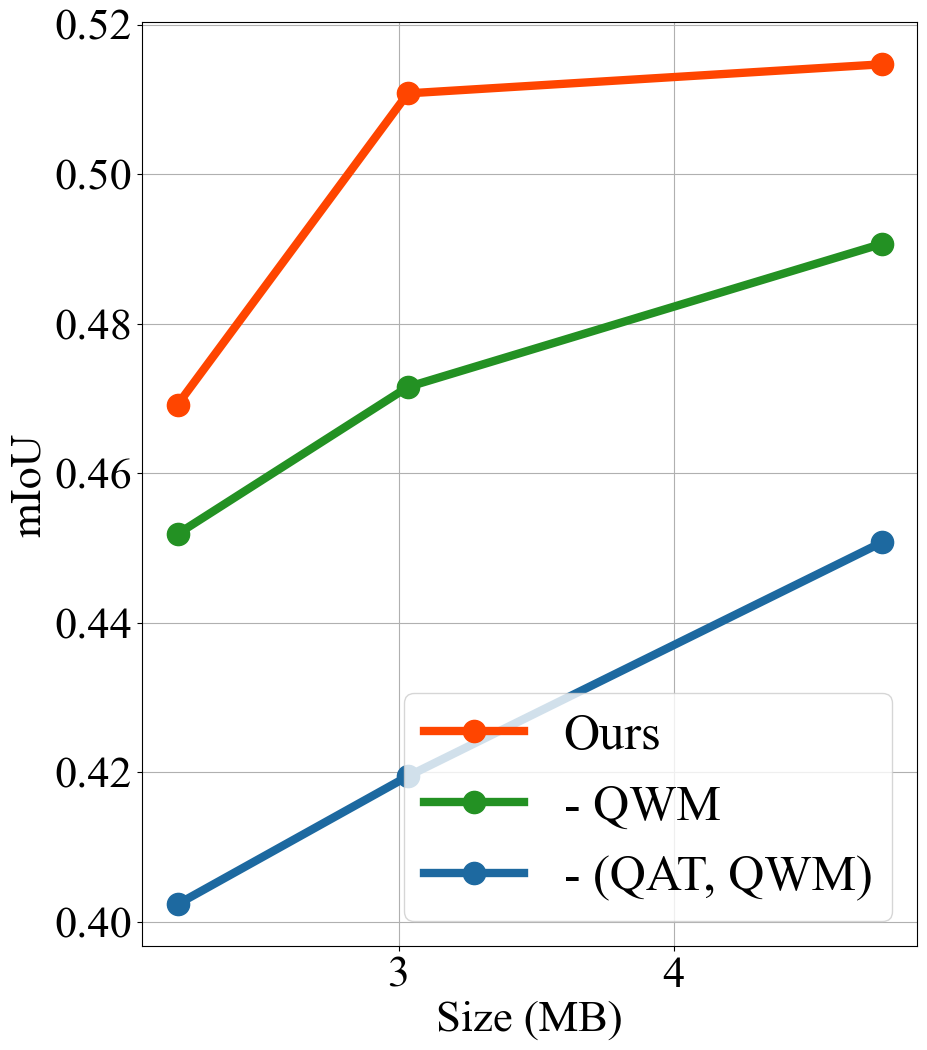}
        \caption{Compression-guided Learning}
        \label{fig:compress-guided}
    \end{subfigure}
    \hfill
    \begin{subfigure}[t]{0.24\textwidth}
        \centering
        \includegraphics[width=\linewidth]{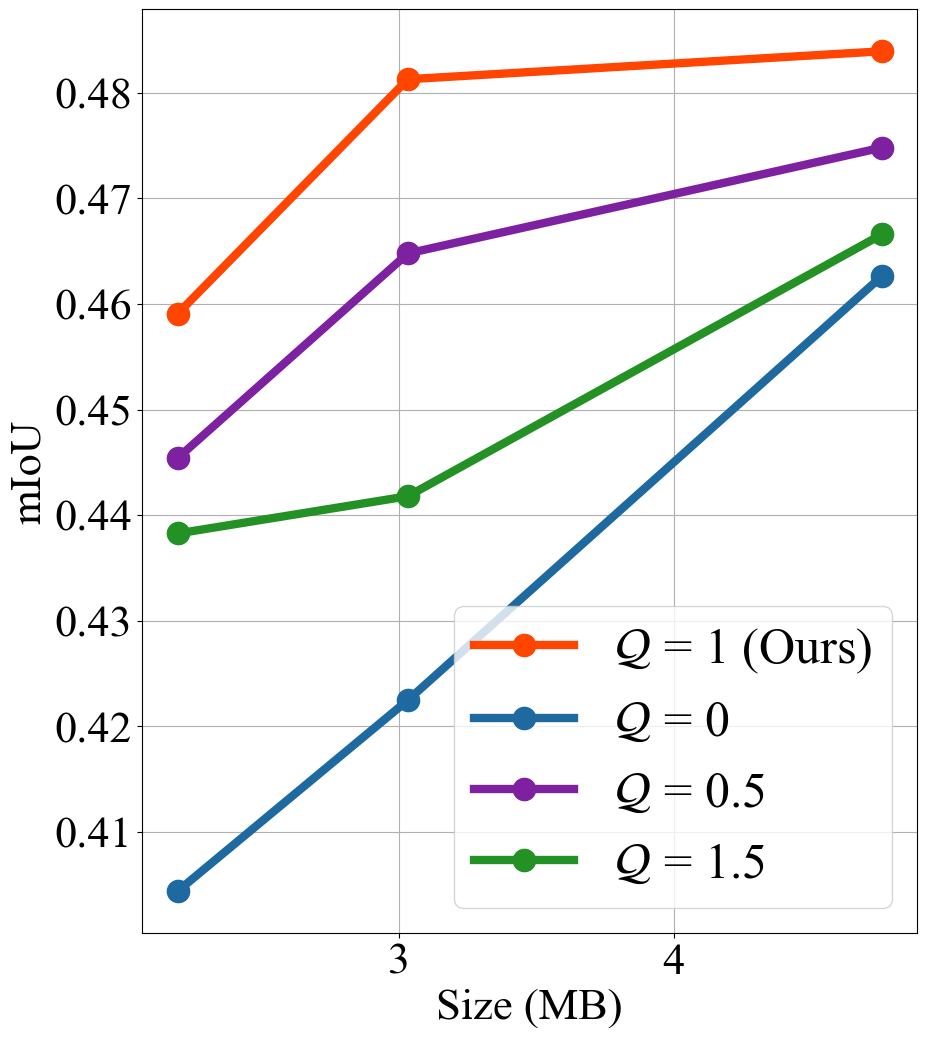}
        \caption{Quantization Step Size}
        \label{fig:rd_qat}
    \end{subfigure}
    \hfill
    \begin{subfigure}[t]{0.24\textwidth}
        \centering
        \includegraphics[width=\linewidth]{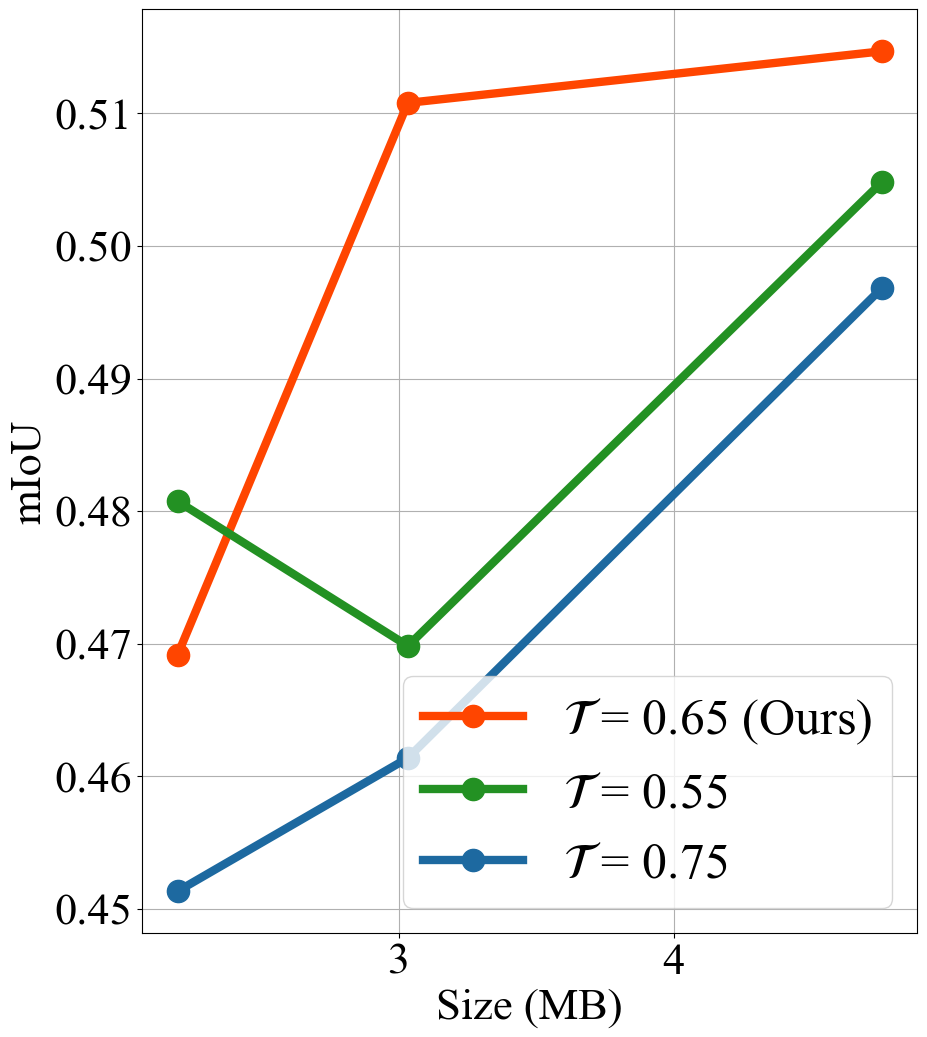}
        \caption{Weighting Threshold}
        \label{fig:rd_thres}
    \end{subfigure}
    
    \caption{Rate-mIoU performance comparison for each ablation study.}
    \label{fig:ablation_rd}
\end{figure*}

  

\vspace{-0.15in}
\paragraph*{\textbf{Comparison with Receiver-side Learning Approach.}}

We present the experimental results for the receiver-side learning scenario depicted in Figure~\ref{fig:teaser}b, where SAM masks are generated from the rendered image for segmentation learning. We compare this approach against our proposed sender-side learning framework. Figure~\ref{fig:rendered} shows that the poor quality of rendered images leads to suboptimal SAM masks and significant failures in segmentation performance, highlighting the necessity of our proposed approach.

\begin{table}[t]
\centering
\caption{Runtime comparison for Stage 1.}
\vspace{-0.5em}
\label{tab:time}
\resizebox{\columnwidth}{!}{%
\begin{tabular}{l|cccc}
\hline
\multirow{2}{*}{\textbf{Method}} & \textbf{Training} & \textbf{Decoding} & {\textbf{Rendering}} \\
 & \textbf{Time (min.)} $\downarrow$ & \textbf{Time (sec.)} $\downarrow$ & \textbf{Speed (FPS)} $\uparrow$ \\
\hline
CAT-3DGS  & 45.8 & 23.1 & 198.9 \\
\hline
Ours & 29.5 & 11.2 & 201.5 \\
\hline
\end{tabular}
}
\vspace{-1em}
\end{table}



\vspace{-0.15in}
\paragraph*{\textbf{Compression-guided Semantic Learning.}}
Quantization-Aware Training (QAT) and Quality-Aware Weighting (QWM) leverage compression information for semantic learning. To assess their contributions, we remove each component one by one and report rate-mIoU performance in Figure~\ref{fig:compress-guided}.
As shown, removing QWM (green curve), which serves to mitigate the negative impact of low-quality Gaussian primitives, leads to a 1-4\% drop in mIoU across bitrates. Further removing QAT (blue curve) causes an additional 5\% drop, confirming its benefits in distinguishing features under contrastive learning. We also provide additional analysis in the supplementary material.

\vspace{-0.15in}
\paragraph*{\textbf{Quantization Step Size.}}

This experiment further examines the impact of applying different quantization step sizes during QAT. We conduct the experiment with QWM disabled to isolate the effects of the quantization alone. In Figure~\ref{fig:rd_qat}, training without QAT ($\mathcal{Q}=0$) yields the lowest performance among all variants. Furthermore, a smaller quantization step size ($\mathcal{Q}=0.5$) limits the potential improvement, while a step size too large ($\mathcal{Q}=1.5$) may introduce excessive noise and harm performance. Our choice of $\mathcal{Q}=1$ strikes a balance and results in superior segmentation accuracy.

\begin{figure}[t]
  \setlength{\abovecaptionskip}{2pt}
  \setlength{\belowcaptionskip}{0pt}
  \centering
  \begin{subfigure}{\linewidth}
    \centering
    \includegraphics[width=\linewidth, clip, trim=1cm 0mm 0mm 11mm]{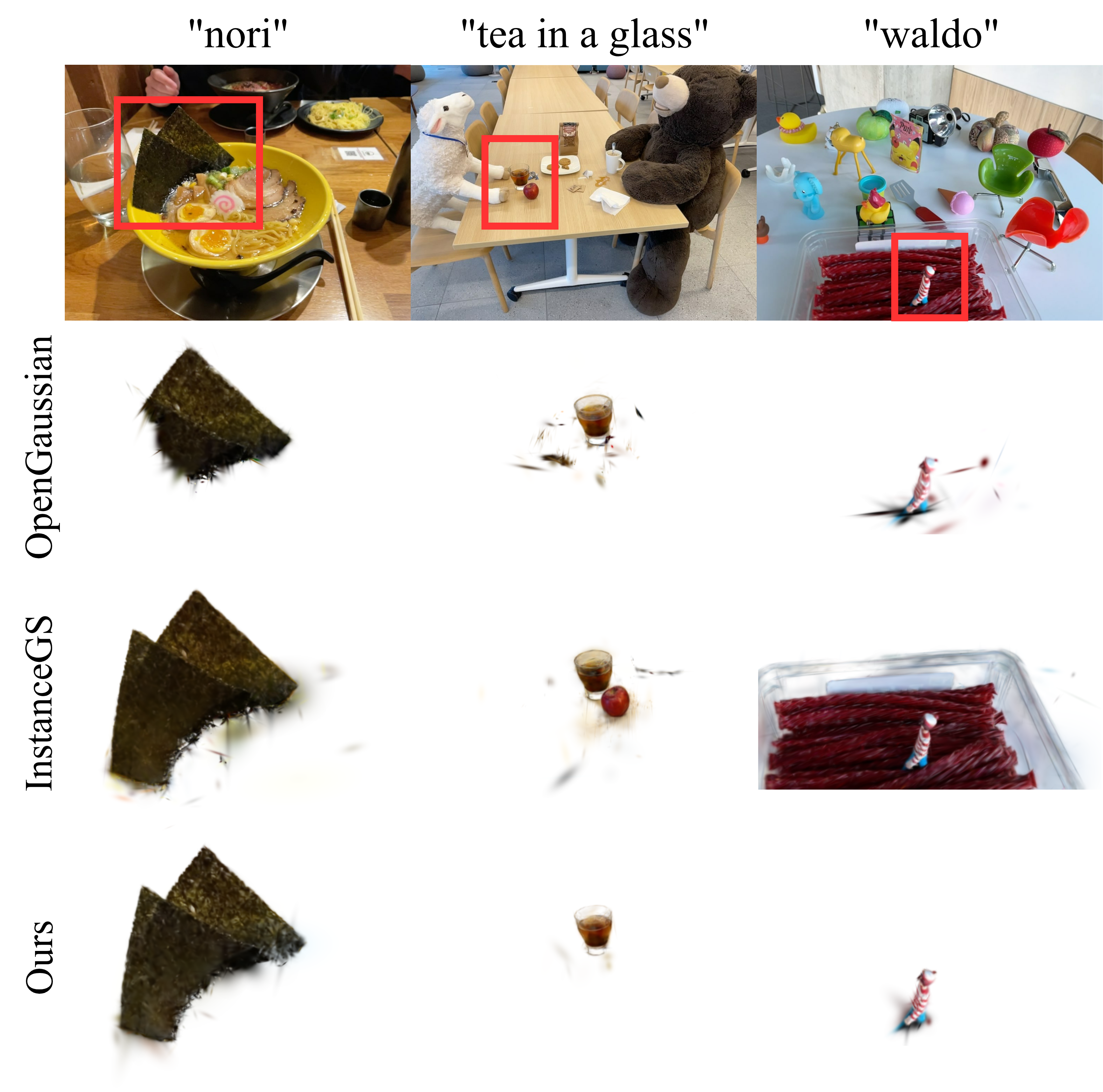}
  \end{subfigure}

  \caption{Qualitative comparison of 3D segmentation on LERF. 
  }
  \vspace{-1em}

  \label{fig:vis}
\end{figure}

\vspace{-0.15in}
\paragraph*{\textbf{Quality-aware Weighting Threshold.}}

Finally, we evaluate the effect of the threshold \( \mathcal{T} \) of QWM, which is used in Eq.~\ref{eq:qwm} to distinguish the high- and low-quality anchors for applying QWM. Figure~\ref{fig:rd_thres} reports the results across different values of \( \mathcal{T} \). 
An overly high threshold ($\mathcal{T}=0.75$) may exclude many valid high-quality anchors, while a threshold too low ($\mathcal{T}=0.55$) insufficiently suppresses low-quality ones. Our chosen threshold of $\mathcal{T}=0.65$ achieves the best overall mIoU performance, validating its effectiveness.  

\subsection{Visualization}
\label{sec:exp/visualization}

A qualitative comparison on open vocabulary segmentation with OpenGaussian~\cite{wu2024opengaussian} and InstanceGS~\cite{li2025instancegaussian} is presented in Figure~\ref{fig:vis}. Our method is able to more effectively separate objects with noticeably clearer boundaries.
In contrast, InstanceGS tends to include neighboring objects within the same segment (see ``waldo"), while OpenGaussian results in incomplete segmentation (see ``nori"). In addition to open-vocabulary segmentation, we present the result for click-based object selection of our proposed framework, similarly to~\cite{wu2024opengaussian, cen2025segment, choi2024click}. 
By clicking on an object of a 2D rendered image, we can successfully retrieve the corresponding 3D object, as shown Figure~\ref{fig:vis_click}. Additional visualizations can be found in the supplementary material.

\begin{figure}[t]
  \setlength{\abovecaptionskip}{0pt}
  \setlength{\belowcaptionskip}{0pt}
  \centering
  \begin{subfigure}{\linewidth}
    \centering
    \includegraphics[width=\linewidth, clip, trim=1cm 0mm 0mm 1mm]{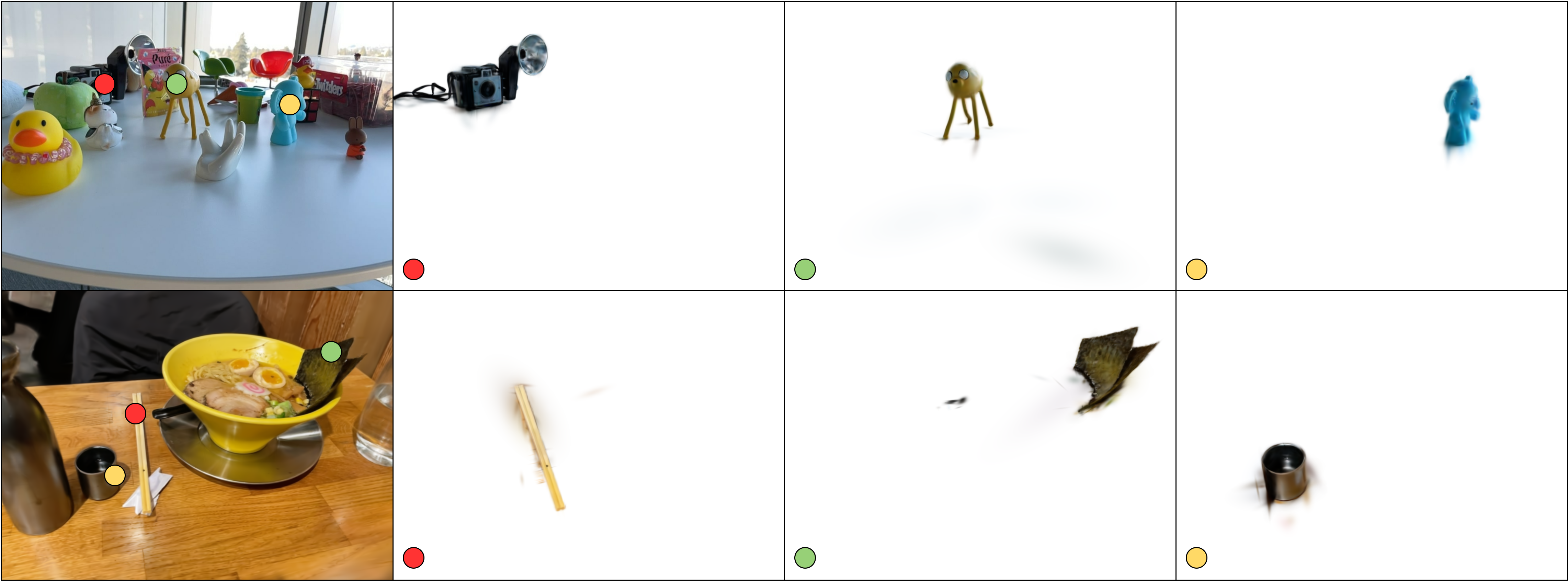}
  \end{subfigure}
  \vspace{1pt} 

  \caption{Click-based Object Selection.}

  \label{fig:vis_click}
\end{figure}

\section{Conclusion}
This work introduces a novel framework for 3DGS that simultaneously addresses rate-distortion compression and segmentation. Our lightweight INR-based hyperprior significantly reduces the size of 3DGS while maintaining high rendering quality and efficient decoding. In addition, the proposed compression-guided segmentation learning further enhances segmentation performance. Together, these contributions enable an efficient and semantically meaningful representation of 3DGS, paving the way for practical downstream applications. We also discuss the limitations of our framework and potential future work in the supplementary material.

\section{Acknowledgement}
This work is supported by MediaTek Advanced Research Center and National Science and Technology Council (NSTC), Taiwan, under Grants 113-2634-F-A49-007-, 112-2221-E-A49-092-MY3, and 114-2221-E-A49-035-MY3. We thank to National Center for High-performance Computing (NCHC) for providing computational and storage resources.
{
    \small
    \bibliographystyle{ieeenat_fullname}
    \bibliography{main}

@String(AAAI = {AAAI})

@inproceedings{kirillov2023segment,
  title={Segment anything},
  author={Kirillov, Alexander and Mintun, Eric and Ravi, Nikhila and Mao, Hanzi and Rolland, Chloe and Gustafson, Laura and Xiao, Tete and Whitehead, Spencer and Berg, Alexander C and Lo, Wan-Yen and others},
  booktitle={IEEE/CVF International Conference on Computer Vision},
  year={2023}
}

@inproceedings{radford2021learning,
  title={Learning transferable visual models from natural language supervision},
  author={Radford, Alec and Kim, Jong Wook and Hallacy, Chris and Ramesh, Aditya and Goh, Gabriel and Agarwal, Sandhini and Sastry, Girish and Askell, Amanda and Mishkin, Pamela and Clark, Jack and others},
  booktitle={International Conference on Machine Learning},
  year={2021},
}

@inproceedings{caron2021emerging,
  title={Emerging properties in self-supervised vision transformers},
  author={Caron, Mathilde and Touvron, Hugo and Misra, Ishan and J{\'e}gou, Herv{\'e} and Mairal, Julien and Bojanowski, Piotr and Joulin, Armand},
  booktitle={IEEE/CVF International Conference on Computer Vision},
  year={2021}
}

@inproceedings{lilanguage,
    title={Language-driven Semantic Segmentation},
    author={Li, Boyi and Weinberger, Kilian Q and Belongie, Serge and Koltun, Vladlen and Ranftl, Rene},
    booktitle={International Conference on Learning Representations},
    year={2022}
}

@article{wu2024opengaussian,
  title={Opengaussian: Towards point-level 3d gaussian-based open vocabulary understanding},
  author={Wu, Yanmin and Meng, Jiarui and Li, Haijie and Wu, Chenming and Shi, Yahao and Cheng, Xinhua and Zhao, Chen and Feng, Haocheng and Ding, Errui and Wang, Jingdong and others},
  journal={Advances in Neural Information Processing Systems},
  year={2024}
}

@inproceedings{li2025instancegaussian,
  title={Instancegaussian: Appearance-semantic joint gaussian representation for 3d instance-level perception},
  author={Li, Haijie and Wu, Yanmin and Meng, Jiarui and Gao, Qiankun and Zhang, Zhiyao and Wang, Ronggang and Zhang, Jian},
  booktitle={IEEE/CVF Computer Vision and Pattern Recognition Conference},
  year={2025}
}

@inproceedings{dai2025efficient,
  title={Efficient Decoupled Feature 3D Gaussian Splatting via Hierarchical Compression},
  author={Dai, Zhenqi and Liu, Ting and Zhang, Yanning},
  booktitle={IEEE/CVF Conference on Computer Vision and Pattern Recognition},
  year={2025}
}

@inproceedings{qin2024langsplat,
  title={Langsplat: 3d language gaussian splatting},
  author={Qin, Minghan and Li, Wanhua and Zhou, Jiawei and Wang, Haoqian and Pfister, Hanspeter},
  booktitle={IEEE/CVF Conference on Computer Vision and Pattern Recognition},
  year={2024}
}

@inproceedings{shi2024language,
  title={Language embedded 3d gaussians for open-vocabulary scene understanding},
  author={Shi, Jin-Chuan and Wang, Miao and Duan, Hao-Bin and Guan, Shao-Hua},
  booktitle={IEEE/CVF Conference on Computer Vision and Pattern Recognition},
  year={2024}
}

@inproceedings{zhou2024feature,
  title={Feature 3dgs: Supercharging 3d gaussian splatting to enable distilled feature fields},
  author={Zhou, Shijie and Chang, Haoran and Jiang, Sicheng and Fan, Zhiwen and Zhu, Zehao and Xu, Dejia and Chari, Pradyumna and You, Suya and Wang, Zhangyang and Kadambi, Achuta},
  booktitle={IEEE/CVF Conference on Computer Vision and Pattern Recognition},
  year={2024}
}

@inproceedings{cen2025segment,
  title={Segment any 3d gaussians},
  author={Cen, Jiazhong and Fang, Jiemin and Yang, Chen and Xie, Lingxi and Zhang, Xiaopeng and Shen, Wei and Tian, Qi},
  booktitle={the AAAI Conference on Artificial Intelligence},
  year={2025}
}

@inproceedings{zhangeconsg,
    title={econSG: Efficient and Multi-view Consistent Open-Vocabulary 3D Semantic Gaussians},
    author={Zhang, Can and Lee, Gim Hee},
    booktitle={International Conference on Learning Representations},
    year={2025}
}

@inproceedings{choi2024click,
  title={Click-gaussian: Interactive segmentation to any 3d gaussians},
  author={Choi, Seokhun and Song, Hyeonseop and Kim, Jaechul and Kim, Taehyeong and Do, Hoseok},
  booktitle={European Conference on Computer Vision},
  year={2024},
}

@inproceedings{
peng2025d,
title={3D Vision-Language Gaussian Splatting},
author={Qucheng Peng and Benjamin Planche and Zhongpai Gao and Meng Zheng and Anwesa Choudhuri and Terrence Chen and Chen Chen and Ziyan Wu},
booktitle={International Conference on Learning Representations},
year={2025},
}

@article{lyu2024gaga,
  title={Gaga: Group Any Gaussians via 3D-aware Memory Bank},
  author={Lyu, Weijie and Li, Xueting and Kundu, Abhijit and Tsai, Yi-Hsuan and Yang, Ming-Hsuan},
  journal={arXiv preprint arXiv:2404.07977},
  year={2024}
}

@inproceedings{ye2024gaussian,
  title={Gaussian grouping: Segment and edit anything in 3d scenes},
  author={Ye, Mingqiao and Danelljan, Martin and Yu, Fisher and Ke, Lei},
  booktitle={European Conference on Computer Vision},
  year={2024},
}

@inproceedings{chen2024gaussianeditor,
  title={Gaussianeditor: Swift and controllable 3d editing with gaussian splatting},
  author={Chen, Yiwen and Chen, Zilong and Zhang, Chi and Wang, Feng and Yang, Xiaofeng and Wang, Yikai and Cai, Zhongang and Yang, Lei and Liu, Huaping and Lin, Guosheng},
  booktitle={IEEE/CVF Conference on Computer Vision and Pattern Recognition},
  year={2024}
}

@inproceedings{huang20253d,
  title={3d gaussian inpainting with depth-guided cross-view consistency},
  author={Huang, Sheng-Yu and Chou, Zi-Ting and Wang, Yu-Chiang Frank},
  booktitle={IEEE/CVF Conference on Computer Vision and Pattern Recognition Conference},
  year={2025}
}

@inproceedings{wu2025aurafusion360,
  title={AuraFusion360: Augmented Unseen Region Alignment for Reference-based 360deg Unbounded Scene Inpainting},
  author={Wu, Chung-Ho and Chen, Yang-Jung and Chen, Ying-Huan and Lee, Jie-Ying and Ke, Bo-Hsu and Mu, Chun-Wei Tuan and Huang, Yi-Chuan and Lin, Chin-Yang and Chen, Min-Hung and Lin, Yen-Yu and others},
  booktitle={IEEE/CVF Conference on Computer Vision and Pattern Recognition Conference},
  year={2025}
}

@inproceedings{xiao2025localized,
  title={Localized gaussian splatting editing with contextual awareness},
  author={Xiao, Hanyuan and Chen, Yingshu and Huang, Huajian and Xiong, Haolin and Yang, Jing and Prasad, Pratusha and Zhao, Yajie},
  booktitle={IEEE/CVF Winter Conference on Applications of Computer Vision},
  year={2025},
}

@inproceedings{kerr2023lerf,
  title={Lerf: Language embedded radiance fields},
  author={Kerr, Justin and Kim, Chung Min and Goldberg, Ken and Kanazawa, Angjoo and Tancik, Matthew},
  booktitle={IEEE/CVF International Conference on Computer Vision},
  year={2023}
}

@article{liu2023weakly,
  title={Weakly supervised 3d open-vocabulary segmentation},
  author={Liu, Kunhao and Zhan, Fangneng and Zhang, Jiahui and Xu, Muyu and Yu, Yingchen and El Saddik, Abdulmotaleb and Theobalt, Christian and Xing, Eric and Lu, Shijian},
  journal={Advances in Neural Information Processing Systems},
  year={2023}
}

@article{yu2024noisynnexploringimpactinformation,
  title={Noisynn: Exploring the influence of information entropy change in learning systems},
  author={Yu, Xiaowei and Huang, Zhe and Xue, Yao and Zhang, Lu and Wang, Li and Liu, Tianming and Zhu, Dajiang},
  journal={arXiv preprint arXiv:2309.10625},
  year={2023}
}

@ARTICLE{10003114,
  author={Li, Xuelong},
  journal={IEEE Transactions on Neural Networks and Learning Systems}, 
  title={Positive-Incentive Noise}, 
  year={2024},
  doi={10.1109/TNNLS.2022.3224577}}

@article{vaswani2017attention,
  title={Attention is all you need},
  author={Vaswani, Ashish and Shazeer, Noam and Parmar, Niki and Uszkoreit, Jakob and Jones, Llion and Gomez, Aidan N and Kaiser, Lukasz and Polosukhin, Illia},
  journal={Advances in neural information processing systems},
  year={2017}
}

@inproceedings{balle2017end,
  title={End-to-end optimized image compression},
  author={Ball{\'e}, Johannes and Laparra, Valero and Simoncelli, Eero P},
  booktitle={International Conference on Learning Representations},
  year={2017}
}

@inproceedings{wang2024end,
  title={End-to-end rate-distortion optimized 3d gaussian representation},
  author={Wang, Henan and Zhu, Hanxin and He, Tianyu and Feng, Runsen and Deng, Jiajun and Bian, Jiang and Chen, Zhibo},
  booktitle={European Conference on Computer Vision},
  year={2024},
}

@inproceedings{chen2024hachashgridassistedcontext,
  title={Hac: Hash-grid assisted context for 3d gaussian splatting compression},
  author={Chen, Yihang and Wu, Qianyi and Lin, Weiyao and Harandi, Mehrtash and Cai, Jianfei},
  booktitle={European Conference on Computer Vision},
  year={2024},
}

@article{chen2025hac++,
    title={Hac++: Towards 100x compression of 3d gaussian splatting},
    author={Chen, Yihang and Wu, Qianyi and Lin, Weiyao and Harandi, Mehrtash and Cai, Jianfei},
    journal={arXiv preprint arXiv:2501.12255},
    year={2025}
}

@inproceedings{lu2023scaffoldgsstructured3dgaussians,
  title={Scaffold-gs: Structured 3d gaussians for view-adaptive rendering},
  author={Lu, Tao and Yu, Mulin and Xu, Linning and Xiangli, Yuanbo and Wang, Limin and Lin, Dahua and Dai, Bo},
  booktitle={IEEE/CVF Conference on Computer Vision and Pattern Recognition},
  year={2024}
}

@inproceedings{zhan2025cat3dgscontextadaptivetriplaneapproach,
    title={CAT-3DGS: A Context-Adaptive Triplane Approach to Rate-Distortion-Optimized 3DGS Compression},
    author={Zhan, Yu-Ting and Ho, Cheng-Yuan and Yang, Hebi and Chen, Yi-Hsin and Chiang, Jui Chiu and Liu, Yu-Lun and Peng, Wen-Hsiao},
    booktitle={International Conference on Learning Representations},
    year={2025}
}

@article{kerbl20233dgaussiansplattingrealtime,
  title={3D Gaussian splatting for real-time radiance field rendering.},
  author={Kerbl, Bernhard and Kopanas, Georgios and Leimk{\"u}hler, Thomas and Drettakis, George},
  journal={ACM Transactions on Graphics},
  year={2023}
}

@article{wang2024contextgscompact3dgaussian,
  title={Contextgs: Compact 3d gaussian splatting with anchor level context model},
  author={Wang, Yufei and Li, Zhihao and Guo, Lanqing and Yang, Wenhan and Kot, Alex and Wen, Bihan},
  journal={Advances in neural information processing systems},
  year={2024}
}

@inproceedings{lee2024compact3dgaussianrepresentation,
  title={Compact 3d gaussian representation for radiance field},
  author={Lee, Joo Chan and Rho, Daniel and Sun, Xiangyu and Ko, Jong Hwan and Park, Eunbyung},
  booktitle={IEEE/CVF Conference on Computer Vision and Pattern Recognition},
  year={2024}
}

@inproceedings{liu2024compgsefficient3dscene,
  title={Compgs: Efficient 3d scene representation via compressed gaussian splatting},
  author={Liu, Xiangrui and Wu, Xinju and Zhang, Pingping and Wang, Shiqi and Li, Zhu and Kwong, Sam},
  booktitle={ACM International Conference on Multimedia},
  year={2024}
}

@inproceedings{ballé2018variationalimagecompressionscale,
  title={Variational image compression with a scale hyperprior},
  author={Ball{\'e}, Johannes and Minnen, David and Singh, Saurabh and Hwang, Sung Jin and Johnston, Nick},
  booktitle={International Conference on Learning Representations},
  year={2018}
}

@inproceedings{he2022elicefficientlearnedimage,
  title={Elic: Efficient learned image compression with unevenly grouped space-channel contextual adaptive coding},
  author={He, Dailan and Yang, Ziming and Peng, Weikun and Ma, Rui and Qin, Hongwei and Wang, Yan},
  booktitle={IEEE/CVF Conference on Computer Vision and Pattern Recognition},
  year={2022}
}

@article{liu2025hemgshybridentropymodel,
  title={Hemgs: A hybrid entropy model for 3d gaussian splatting data compression},
  author={Liu, Lei and Chen, Zhenghao and Jiang, Wei and Wang, Wei and Xu, Dong},
  journal={arXiv preprint arXiv:2411.18473},
  year={2024}
}

@article{zhan2025cat3dgspronewbenchmark,
  title={CAT-3DGS Pro: A New Benchmark for Efficient 3DGS Compression},
  author={Zhan, Yu-Ting and Yang, He-bi and Ho, Cheng-Yuan and Chiang, Jui-Chiu and Peng, Wen-Hsiao},
  journal={arXiv preprint arXiv:2503.12862},
  year={2025}
}

@inproceedings{girish2024eaglesefficientaccelerated3d,
  title={Eagles: Efficient accelerated 3d gaussians with lightweight encodings},
  author={Girish, Sharath and Gupta, Kamal and Shrivastava, Abhinav},
  booktitle={European Conference on Computer Vision},
  year={2024},
}

@article{fan2024lightgaussianunbounded3dgaussian,
  title={Lightgaussian: Unbounded 3d gaussian compression with 15x reduction and 200+ fps},
  author={Fan, Zhiwen and Wang, Kevin and Wen, Kairun and Zhu, Zehao and Xu, Dejia and Wang, Zhangyang and others},
  journal={Advances in neural information processing systems},
  year={2024}
}

@inproceedings{ali2024trimmingfatefficientcompression,
author    = {Muhammad Salman Ali and Maryam Qamar and Sung-Ho Bae and Enzo Tartaglione},
title     = {Trimming the Fat: Efficient Compression of 3D Gaussian Splats through Pruning},
booktitle = {British Machine Vision Conference},
year      = {2024},
}

@article{ren2024octreegsconsistentrealtimerendering,
  title={Octree-GS: Towards Consistent Real-time Rendering with LOD-Structured 3D Gaussians},
  author={Ren, Kerui and Jiang, Lihan and Lu, Tao and Yu, Mulin and Xu, Linning and Ni, Zhangkai and Dai, Bo},
  journal={IEEE Transactions on Pattern Analysis and Machine Intelligence},
  year={2025},
}

@inproceedings{sun2024f3dgsfactorizedcoordinatesrepresentations,
  title={F-3dgs: Factorized coordinates and representations for 3d gaussian splatting},
  author={Sun, Xiangyu and Lee, Joo Chan and Rho, Daniel and Ko, Jong Hwan and Ali, Usman and Park, Eunbyung},
  booktitle={ACM International Conference on Multimedia},
  year={2024}
}

@inproceedings{navaneet2024compgssmallerfastergaussian,
  title={Compgs: Smaller and faster gaussian splatting with vector quantization},
  author={Navaneet, KL and Pourahmadi Meibodi, Kossar and Abbasi Koohpayegani, Soroush and Pirsiavash, Hamed},
  booktitle={European Conference on Computer Vision},
  year={2024},
}

@inproceedings{niedermayr2023compressed,
  title={Compressed 3d gaussian splatting for accelerated novel view synthesis},
  author={Niedermayr, Simon and Stumpfegger, Josef and Westermann, R{\"u}diger},
  booktitle={IEEE/CVF Conference on Computer Vision and Pattern Recognition},
  year={2024}
}
}

\clearpage        

\appendix
\newpage


\end{document}